\pdfoutput=1

\documentclass[11pt]{article}

\usepackage[]{EMNLP2022}

\usepackage{times}
\usepackage{latexsym}

\usepackage[T1]{fontenc}

\usepackage[utf8]{inputenc}

\usepackage{microtype}

\usepackage{inconsolata}

\usepackage{graphicx}
\usepackage{multirow}
\usepackage{booktabs}

\newcommand{\ourmodel}{{\textsc{FactEdit}}}
\newcommand{\Sref}[1]{\S\ref{#1}}
\newcommand{\fref}[1]{Figure~\ref{#1}}

\newcommand{\Tref}[1]{Table~\ref{#1}}

\title{Correcting Diverse Factual Errors in Abstractive Summarization via Post-Editing and Language Model Infilling}

\author{Vidhisha Balachandran$^{\clubsuit}$ \quad Hannaneh Hajishirzi$^{\diamondsuit \heartsuit}$ \\ \quad \textbf{William W. Cohen}$^\spadesuit$ \quad \textbf{Yulia Tsvetkov}$^\heartsuit$ \\
$^\clubsuit$Language Technologies Institute, Carnegie Mellon University\\
$^\diamondsuit$Allen Institute for Artificial Intelligence \\
 $^\heartsuit$Paul G.~Allen School of Computer Science \& Engineering, University of Washington \\
 $^\spadesuit$Google Research \\
\texttt{\small vbalacha@cs.cmu.edu, hannaneh@cs.washington.edu, wcohen@google.com, yuliats@cs.washington.edu}}

\begin{document}
\maketitle

\begin{abstract}
Abstractive summarization models often generate inconsistent summaries containing factual errors or hallucinated content. 
Recent works focus on correcting factual errors in generated summaries via post-editing. Such correction models are trained using adversarial non-factual summaries constructed using heuristic rules for injecting errors.
However, generating non-factual summaries using heuristics often does not generalize well to actual model errors. 
In this work, we propose to generate hard, representative synthetic examples of non-factual summaries through infilling language models. With this data, we train a more robust fact-correction model to post-edit the summaries to improve factual consistency. Through quantitative and qualitative experiments on two popular summarization datasets--- CNN/DM and XSum---we show that our approach vastly outperforms prior methods in correcting  erroneous summaries. Our model---\ourmodel{}---improves factuality scores by over $\sim$11 points 
on CNN/DM and over $\sim$31 points on XSum on average across multiple summarization models, producing more factual summaries while maintaining competitive summarization quality.\footnote{Code and data available at \url{https://github.com/vidhishanair/FactEdit}.}

\end{abstract}

\section{Introduction}
While modern summarization models generate highly fluent summaries that appear realistic~\citep{bart, zhang2020pegasus}, these models are prone to generating non-factual and sometimes entirely fabricated content \cite{fact-aware, goodrich19, factentailment}. 
With the increasing adoption of language generation tools in user-facing products, such unreliability poses severe risks, including the spread of misinformation, panic and other potentially harmful effects \cite{aigenfakenews, hutson2021robo}.
\begin{figure}[t]
    {\includegraphics[width=\columnwidth]{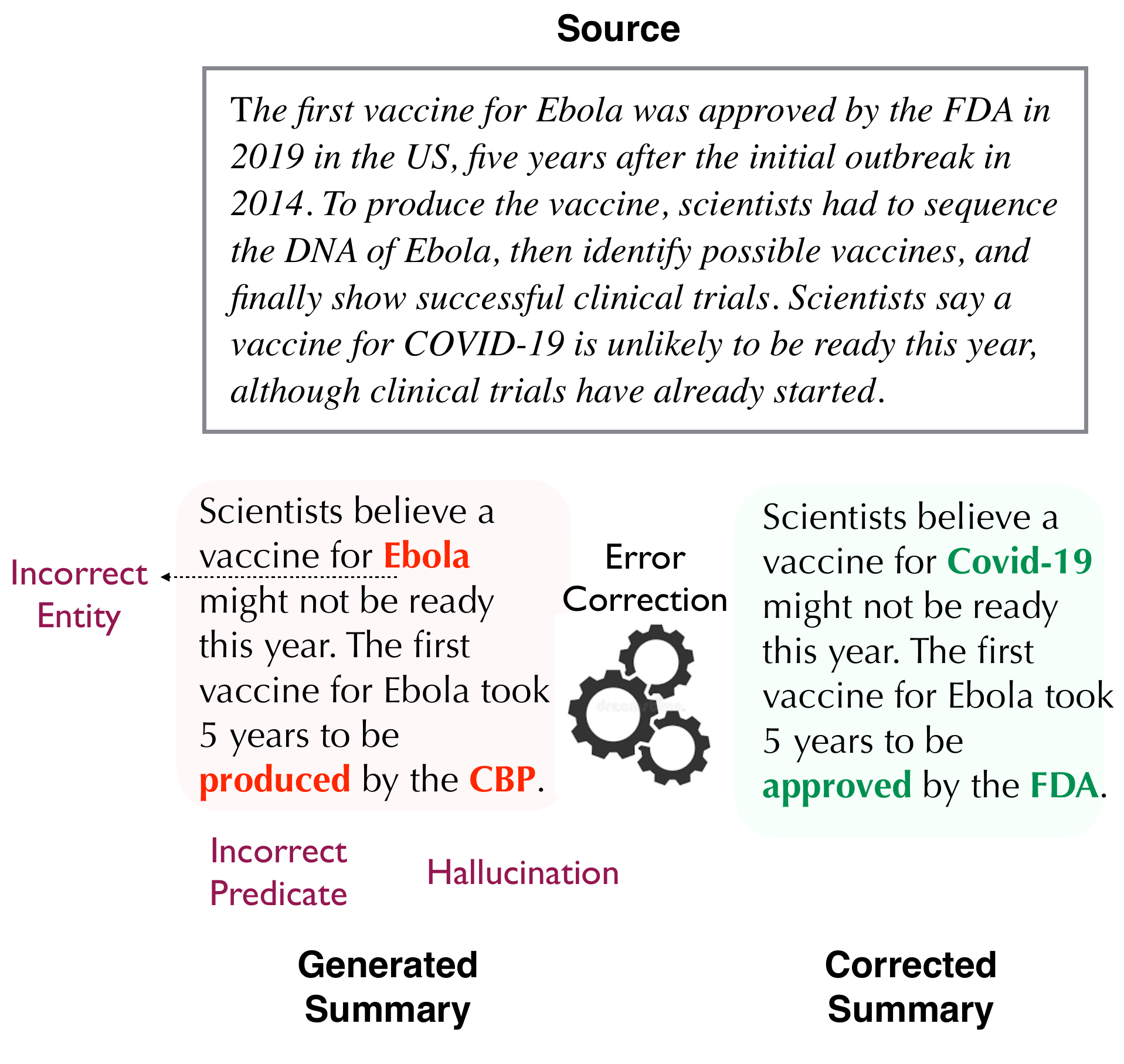}}
    \caption{Model generated summaries often produce content which is factually inconsistent w.r.t.~to the source. \ourmodel{} rewrites these summaries by maintaining the abstractiveness but correcting factual errors.}
    \label{fig:corr_example}
\end{figure}
\begin{figure*}[t]
    {\includegraphics[width=0.96\textwidth]{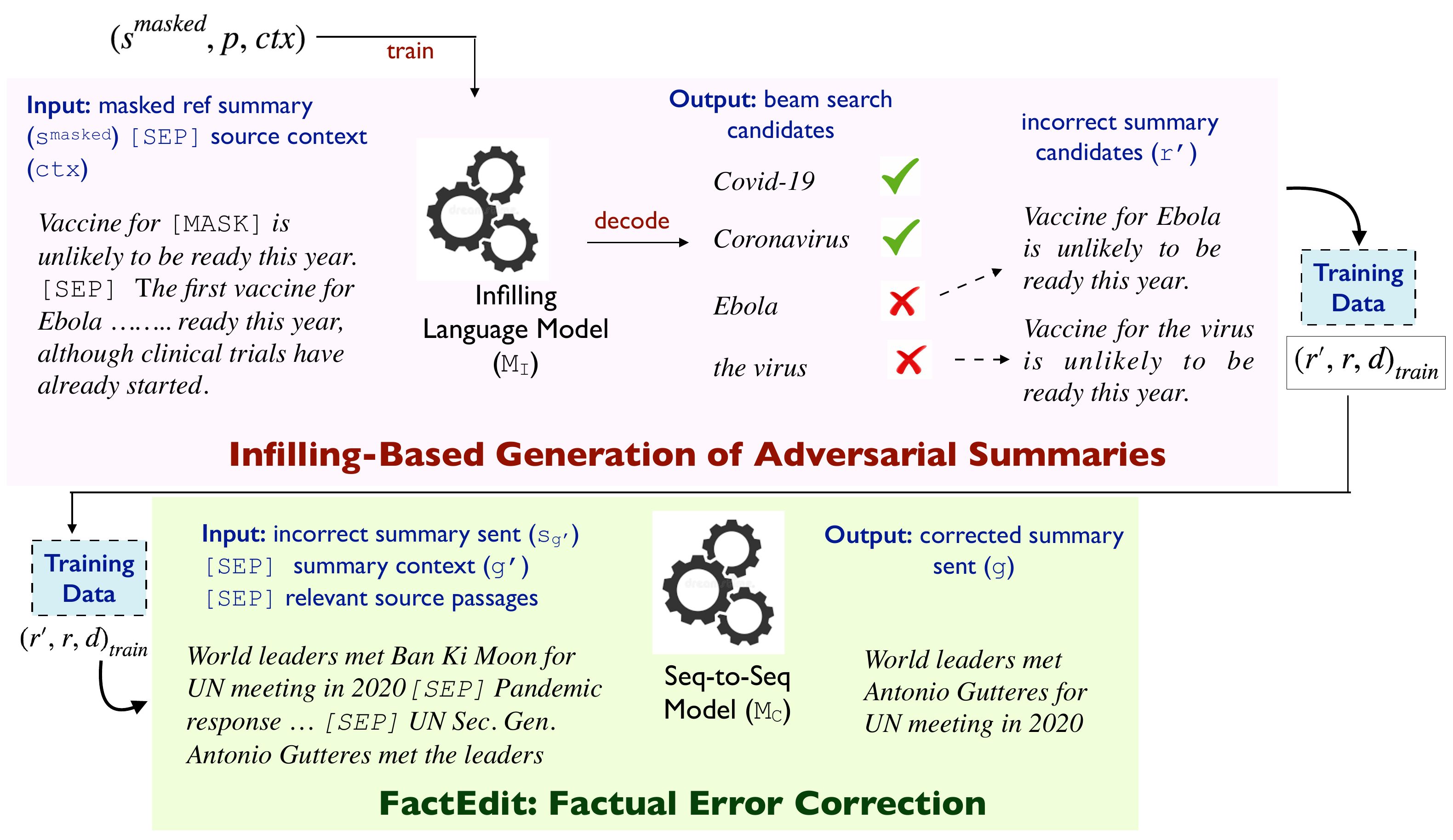}}
    \caption{Architecture framework for \ourmodel{}. Using masked versions of existing reference summaries, we use an infilling language model to produce alternative candidates for the mask position. We construct factually incorrect summaries by replacing the mask with the lower ranked candidates. Finally, we train a sequence-to-sequence model for fact correction using the synthetically constructed data.}
    \label{fig:overview}
\vspace{-0.3cm}
\end{figure*}

Since it is difficult to control for factuality at training or inference time \cite{huang2021factual, dreyer2021analyzing}, a popular approach to fix the factual inconsistencies is via post-editing generated summaries \cite{cao-etal-2020-factual, dong-etal-2020-multi}. This allows summarization models to focus on fluency and content-relevance while improving factual consistency.
However, there is no suitable data for training post-editing models to directly ``translate'' an incorrect summary to a correct one. 
Prior work constructed synthetic training data by introducing simple heuristic errors like replacing entities or numbers in reference summaries \cite{cao-etal-2020-factual}, but it is not clear whether such synthetic errors have sufficient coverage and accurately represent the types and distribution of actual errors made by language models. 
Further, with increasing language generation capabilities, models make more complex factual errors involving discourse structures and paraphrasing which cannot be easily captured with heuristics \cite{frank}. The goal of our work is to develop post-editing models that generalize over a wider range of factual errors (example in \fref{fig:corr_example}) in generated summaries from diverse summarization model types.


We propose \ourmodel{}---a novel approach to post-editing text, to control for content factuality in generated summaries. 
Rather than manually defining a list of heuristic errors, it incorporates a new algorithm to generate adversarial (non-factual) examples using infilling language models \cite{donahue2020enabling}. We use lower ranked beam-search candidates from the language model as a source for potentially factually-incorrect summary facts, thereby producing a set of plausible, likely, and fluent, incorrect synthetic summaries for a particular correct reference summary. In this way, we leverage the capabilities of large language models to produce multiple candidates of alternative, erroneous summaries. 
These examples, along with factually correct references, are then used to train a sequence-to-sequence fact-correction model that aims at generating a factually consistent version of the candidate summary (\Sref{sec:model}).

We evaluate \ourmodel{} on two datasets - CNN/DailyMail \cite{cnn-dm} and XSum \cite{xsum} and across nine summarization models with the FRANK benchmark \cite{frank} for evaluating various categories of factual errors in generated summaries (\Sref{sec:expts}). The two summarization datasets represent varied distributions of factual errors in models trained on them and hence constitute a good test bed to evaluate the generalizability of our model.
We show that \ourmodel{} substantially improves factuality scores across two metrics - Ent-DAE \cite{entdae} and FactCC \cite{factcc}. On the Ent-DAE metric, \ourmodel{} improves results by $\sim$11 points 
(CNN/DM) and $\sim$31 points (XSum), and on the FactCC metric we show improvements of $\sim$6 points (CNN/DM) and $\sim$24 (XSum) points on average across models (\Sref{sec:res}).
Further, our analysis  shows that \ourmodel{} effectively corrects diverse error categories without the need for special heuristics or annotations (\Sref{sec:analysis}). An important application of \ourmodel{} is to audit summarization systems and facilitate their reliability.

\section{Model}
\label{sec:model}
Assume a summarization model trained to process a document $d$ and generate a coherent and fluent summary\footnote{We denote incorrect input (to fact correction model) summaries using $'$ and corrected output (from fact correction model) without the $'$ throughout this paper. For E.g: $g'$ is incorrect summary, $r'$ is the incorrect reference summary while $g$ is the corrected summary and $r'$ is the corrected reference summary.} $g'$ which has been shown to often misrepresent facts from the document. \ourmodel{} is a fact correction model $M_C$ which takes the generated summary $g'$ and document $d$, identifies factual errors and generates a rewritten summary $g$ by correcting them (as outlined in \fref{fig:overview}).

We present an adversarial data generation approach which leverages the power of pre-trained language models to produce fluent and complex factually incorrect summaries. We train an infilling language model $\mathbf{M}_I$ using documents from summarization training data and use the model to introduce diverse factual errors in sentences from them (\Sref{sec:model_infill}). Using the trained model, we introduce factual errors in reference summaries of the training data $r$ producing an incorrect summary $r'$ resulting in a synthetic dataset $\{r',r,d\}_{train}$ of erroneous summaries mapped to their corrected versions (pink section in \fref{fig:overview}). We train a sequence-to-sequence model $\mathbf{M}_C$ for factual error correction using the generated synthetic data (\Sref{sec:model_corr}). Finally, we use the trained correction model to rewrite model generated summaries $g'$ producing a corrected version $g$ (\Sref{sec:model_inf} - green section in \fref{fig:overview}).


\subsection{Infilling Data Generator $\mathbf{M}_I$}
\label{sec:model_infill}
Our data generation process leverages infilling language models \cite{donahue2020enabling} to produce candidates to fill masked phrases in a summary sentence. We mask parts of the input and use the infilling model to generate multiple candidates for the masked position. We then use lower order beam candidates as potential incorrect candidates to generate an incorrect version of the input. We hypothesize that, given the relevant context of a source document, a strong language model generates relevant and factual sequences at higher probabilities, compared to lower probability sequences.
For the infilling model, we hypothesize that the lower ranked candidates are often alternative phrases of similar types (in case of entities) or parts-of-speech which are plausible but often not factually correct. Motivated by prior work \cite{dae} using lower ranked beam search candidates as a source for adversarial data, we use the lower ranked candidates to construct erroneous summaries from reference summaries. 

\paragraph{Training:} Our infilling model $\mathbf{M}_I$ is trained to take a masked sentence 
$s^{masked}$ 
and its relevant context $ctx$ as input and generate a correct phrase to fill in the masked span. To train  $\mathbf{M}_I$, we construct a dataset using documents $d$ from the training data of existing summarization datasets. For each sentence $s$ in the first-$k$ ($k$=5) positional sentences of a document $d$, we identify the subjects, objects and relations \{sub,~obj,~rel\} in them using OpenIE \cite{openie}. By iteratively masking each phrase $p$ in \{sub,obj,rel\}, we create a masked query 
$s^{masked}$ 
and its corresponding context $ctx$ by removing the masked sentence from the document, resulting in our training data
$\{s^{masked}, p, ctx\}$, 
where $p$ is the masked span text. We train a sequence-to-sequence model $\mathbf{M}_I$ on this data which takes
\textit{$s^{masked}$ \texttt{[SEP]} $ctx$}
as input and learns to generate $p$ as the output. We intentionally use only sentences from the document as masked queries and do not use sentences from the reference summaries, to ensure that the model does not memorize phrases from the references. Thus, when applied to unseen reference sentences during inference, the model will produces richer beam search candidates.


\paragraph{Adversarial Data Generation:} We use the trained infilling model to generate the synthetic dataset for fact correction using the document reference pairs $\{d,r\}_{train}$ from the summarization training data. For each sentence in the reference $s_r$, we use OpenIE to extract \{sub,~obj,~rel\} and iteratively mask one phrase at a time to construct 
masked sentences $s^{masked}$ from the references. 
We provide this masked reference summary sentence 
and document $d$ as input to the model and perform beam-search decoding for generation.
We then consider lower ranked beam candidates (rank=[5,15])\footnote{We chose this range of ranks based on a manual analysis of 500 generated adversarial examples where our method produced factually incorrect replacements over 90\% of the time.} as non-factual alternatives for the corresponding masked phrase. We then use these candidates as the replacements for the mask producing an erroneous summary $r'$. Running this on the $\{d,r\}_{train}$ training data, we construct a synthetic data $\{r',r,d\}_{train}$ of factually incorrect summaries paired with their correct version where $r'$ and $r$ differ by an incorrect phrase. To train the model to not perform any corrections on factual summaries, we keep original reference summaries for 20\% of the data points ($r'=r$).

\subsection{Fact Correction Model $\mathbf{M}_C$}
\label{sec:model_corr}
Using the parallel data $\{r',r,d\}_{train}$ produced by the above infilling method, we train models for factual error correction. In contrast to prior work which used pointer based models to copy phrases from the source document, we use a sequence-to-sequence model like BART \cite{bart} to preserve the abstractive content in the input. 
The model $\mathbf{M}_C$ is trained with an erroneous reference summary sentence $s_{r'}$ produced by the infilling data generator and the corresponding document $d$ as input and the correct reference summary sentence $s_r$ as output. 
A straightforward option is to provide $s_{r'}, d$ concatenated as inputs to the model. But we hypothesize that providing the right context can help the model better correct the errors. Below we outline input structures that provide better context in the input: \\
\textbf{Relevant Supporting Passages: } To help the model better connect the relevant facts in the source document to the summary sentence being corrected, we experiment with providing only the most relevant parts of the document as input context instead of the entire document. Using a scoring function (ROUGE), we identify sentences from the document which have high overlap with the generated summary sentence and extract the top-$k$ ($k$=3 for our work) such sentences. We provide these sentences along with a window of $w_k$ ($w_k$=2) sentences before and after each as the input context to the model.\\
\noindent\textbf{Surrounding Summary Context: } While simple errors like incorrect entities can be detected and corrected with only the context of the current sentence being corrected, more complex discourse level errors like incorrect pronouns require the context of the rest of the sentences of the summary. To enable this, we additionally give the complete generated summary (other sentences from the summary) as additional context. For single sentence summaries like headline generation, this does not change the original setting, but for longer summaries this setting helps with discourse level errors.

\noindent  In essence, our model $\mathbf{M}_C$ takes the input as
\textit{Incorrect Reference Sentence ($s_{r'})$ \texttt{[SEP]} Full Reference Summary ($r'$) \texttt{[SEP]} Relevant Passages} and generates the corrected summary $r$ as output. 

\subsection{Inference}
\label{sec:model_inf}
Our trained fact correction model $\mathbf{M}_C$ can be directly applied to any model-generated summaries $g'$, without access to the underlying model. 
For each sentence in a generated summary, we identify the relevant passages using ROUGE and provide it as an input to the model (in the form \textit{Generated Summary Sentence ($s_{g'}$) \texttt{[SEP]} Generated Full Summary ($g'$) \texttt{[SEP]} Relevant Passages}). 


\section{Experiments and Data}
\label{sec:expts}
\subsection{Datasets}
\label{sec:expts_datasets}
We use two news summarization datasets CNN-DailyMail \cite{cnn-dm} and XSum \cite{xsum}. The two datasets have been extensively studied for the factual consistency in their generated summaries across a variety of models \cite{goodrich19, fact-aware}. Reference summaries from CNN/DM are longer, having on average three sentences, and more extractive in nature. XSum on the other hand has shorter, single sentence summaries and is significantly more abstractive in nature. The summaries in these datasets are qualitatively different, and hence models trained on the two datasets present varied levels of challenges in maintaining factual consistencies. 

Prior work have studied summaries generated from different language models and characterized the factual errors in them \citep{frank}.
Generated summaries on the CNN/DM dataset are more extractive in nature and hence are more factual ($\sim$70\% of summaries are factual) than the more abstractive generated summaries of XSum ($\sim$20\% of summaries are factual). 
The longer summaries in CNN/DM display discourse level inconsistencies while summaries from XSum often hallucinate content which is not supported by the source document.
Hence, the two datasets present a varied setting for evaluating the efficacy of our model across different kinds of errors. For our main evaluation, we evaluate the overall capability of \ourmodel{} in correcting errors in summaries generated by a BART model. 

We further evaluate our model on the FRANK benchmark \cite{frank} which contains generated summaries obtained using multiple language models for both datasets annotated with human judgements on their factuality and the category of factual error. As different language models have different distribution of factual error types, this evaluation helps us study the generalizability of \ourmodel{} in correcting errors across them.\footnote{As the benchmark has publicly available model outputs, the summaries across different datasets are from different models owing to their availability.} For the CNN/DM dataset, it contains model outputs from a LSTM Seq-to-Seq model (S2S) \citep{rush}, a Pointer-Generator Network (PGN) model \citep{pgn}, a Bottom-Up Summarization (BUS) model \citep{bus}, a Bert based Extractive-Abstractive model (BertSum) \citep{liu-lapata-2019-text} and a jointly pretrained transformer based encoder-decoder model BART \citep{bart}. For the XSum dataset, it contains model outputs from a Topic-Aware CNN Model \citep{xsum}, a Pointer-Generator Network (PGN) model, a randomly initialized (TransS2S) \citep{vaswani2017attention} and one initialized with Bert-Base (BertS2S) \citep{bert}. 
\begin{table*}[t]
\centering 
\begin{tabular}{l|l|c|c|c|c|c}
\textbf{Dataset} & \textbf{Method} & \textbf{R1} & \textbf{R2} & \textbf{RL} & \textbf{FactCC} & \textbf{Ent-DAE}\\
\hline
\multirow{3}{*}{CNN/DM} & Bart \cite{bart} & 44.07 & 21.08 & 41.01 & 75.78 & 74.85 \\
 & \citet{cao-etal-2020-factual} & 42.72 & 20.59& 39.92&49.98&74.83 \\
 & \ourmodel{} & 42.17 & 20.22 & 39.37 & 75.49 & \textbf{75.71} \\
  & \ourmodel{} + FactCC Filter (FF) & 42.53 & 20.48 & 39.74 & \textbf{76.03} & 75.36 \\
\hline
\multirow{3}{*}{XSum} & Bart \cite{bart} & 34.71 & 15.04 & 27.40 & 21.93 & 20.03 \\
 & \citet{cao-etal-2020-factual} & 33.64 & 14.71& 26.49& 7.01& 20.03\\
 & \ourmodel{} & 33.58 & 14.68 & 26.71 & \textbf{23.91} & \textbf{20.13}\\
 & \ourmodel{} + FactCC Filter (FF) & 33.58 & 14.68 & 26.71 & \textbf{23.91}  &\textbf{20.13} \\

\end{tabular}
\caption{\ourmodel{} performance for correcting BART outputs (best performance in bold). \ourmodel{} ourperforms factuality related baselines on FactCC and DAE scores, while maintaining competitive summarization quality.}
\label{tab:overall_res}
\end{table*}
\subsection{Experimental Settings and Evaluation}
\label{sec:expts_eval}
\noindent\textbf{Setup:} We use OpenIE \cite{openie} to pre-process each summary and extract {subject, object, predicate} triples for each summary sentence.
We use BART-base \cite{bart} as our sequence-to-sequence model for the infilling based data generator and the fact correction model. Both models were trained with a batch size of 48, a learning rate of 3e-5, and warm-up of 1000 for 1 epoch. 
The maximum input sequence length was 512 and maximum output sequence length was 128. Using the infilling data generator, we generate 1233329 negative, 308332 positive examples for CNN/DM and 724304 negative, 181076 positive, examples for XSum as training data for fact correction. Models were trained on 4 Nvidia GeForce GTX TITAN X GPUs and each training run took $\sim$15 hours. All hyperparameters were chosen based on generated dev set ROUGE-L \cite{lin2004rouge} on each dataset.\\
\noindent\textbf{Evaluation Setup:} Evaluating factual consistency of generated summaries is challenging, with relatively recent metrics developed to detect it. These metrics unfortunately do not correlate highly with human judgements yet.
We therefore evaluate our model using two metrics - FactCC \cite{factcc} and Ent-DAE \cite{entdae}; each captures different error types.
FactCC is a binary classifier, trained on a synthetic, heuristic error dataset, which is
better at detecting simple semantic errors like incorrect entities or numbers. Ent-DAE is a classifier trained on synthetic data constructed using the dependency structure of the text. In addition to semantic errors, it is better at detecting more complex discourse-level errors \cite{frank}. 
We also report ROUGE \cite{lin2004rouge} to evaluate if our model maintains the fluency of summaries. While ROUGE is less correlated with factuality \cite{frank, factentailment}, 
it helps evaluate if the corrected summary is fluent and aligned with the reference summary. However, with factual corrections of outputs we expect small drops in ROUGE, since generation models were specifically optimized to maximize ROUGE presumably at the expense of factuality.


Our evaluation has two settings: i) \ourmodel{} - correct all generated summaries in the test set and ii) \ourmodel{} + FactCC Filter (FF) - using the FactCC metric we identify factually incorrect summaries, and only correct the incorrect ones.

\noindent\textbf{Baselines:} We compare our approach with \cite{cao-etal-2020-factual} as the baseline. The baseline uses a heuristic set of rules proposed by \citet{factcc} to introduce simple errors (Entity, Number, Date, and Pronoun) in reference summaries and trains a BART-base model for error correction.
Comparing our model with \cite{cao-etal-2020-factual} helps us evaluate the benefit of our Infilling LM based adversarial data generator. \footnote{While \cite{dong-etal-2020-multi} is also a factual error correction method, we were unable to reproduce it as no public code was available.}

\section{Results}
\label{sec:res}
\subsection{Factuality Results}
\label{sec:res_fact}
We first evaluate \ourmodel{}'s ability to correct errors in summaries generated by a BART-base summarization model on the entire test set. We first generate summaries using a BART-base model finetuned on each dataset and then provide the generated summaries and their corresponding source documents as inputs to \ourmodel{} for correction.
\begin{table}[t]
\centering
\small
\begin{tabular}{l|c|c|c}
\textbf{Method} & \textbf{RL} & \textbf{FactCC} & \textbf{Ent-DAE}\\
\hline
\multicolumn{4}{c}{CNN/DM}\\
\hline
Bart & 41.53 & 46.29 & 72.57 \\
\ourmodel{} & 37.73 & 42.29 & 78.86\\
\ourmodel{} (FF) & 37.73 & \textbf{53.14} & \textbf{81.71}
 \\
 \cline{1-4}
BertSum & 38.74 & 58.86 & 82.29 \\
\ourmodel{} & 35.6 & 55.43 & 79.43\\
\ourmodel{} (FF) & 35.6 &	\textbf{61.71} & \textbf{82.86}\\
 \cline{1-4}
BUS & 38.59 & 49.71 & 70.28\\
\ourmodel{} & 33.79& 48.00 & 76.00\\
\ourmodel{} (FF) & 33.79 & \textbf{56.57} & \textbf{80.00}\\
 \cline{1-4}
PointGen & 35.62 & 80.57 & 93.14\\
\ourmodel{} & 32.54 & 75.43 & 90.29 \\
\ourmodel{} (FF) & 32.54 & 78.29 &	90.86 \\
 \cline{1-4}
Seq2Seq & 27.15 & 19.43 & 29.71 \\
\ourmodel{} & 24.78 & 23.43 & 48.00\\
\ourmodel{} (FF) & 24.78 & \textbf{24.00} & \textbf{54.29}\\
\hline
\multicolumn{4}{c}{XSum}\\
\hline
BertS2S & 29.05 & 22.29 & 05.71 \\
\ourmodel{} & 28.93 & 50.43 & 40.00 \\
\ourmodel{} (FF) & 28.95 & \textbf{50.43} & \textbf{40.00}\\
 \cline{1-4}
TConvS2S & 25.69 & 17.71 & 04.00 \\
\ourmodel{} & 25.64 & 47.16 &	29.14 \\
\ourmodel{} (FF) & 25.64 &	\textbf{47.16} & \textbf{29.14} \\
 \cline{1-4}
PointGen & 23.12 &	18.29 &	00.57 \\
\ourmodel{} & 23.02 & 43.75 & 32.00 \\
\ourmodel{} (FF) & 23.04 & \textbf{43.75}&	\textbf{32.00}\\
 \cline{1-4}
TranS2S & 23.93 & 18.86 & 2.86 \\
\ourmodel{} & 23.86 & 31.73 & 36.00 \\
\ourmodel{} (FF) & 23.86 & \textbf{31.73} & \textbf{36.00} \\
\end{tabular}
\caption{Performance of \ourmodel{} across different model generated summaries in the FRANK setting. Best performance is indicated in Bold. \ourmodel{} model vastly improves factuality across multiple models on both FactCC and DAE scores.
}
\label{tab:frank_res}
\vspace{-1.3em}
\end{table}

Table \ref{tab:overall_res} shows results for this experiment. Our results show that correcting factual errors using our model improves the factuality results. The baseline model performs poorly with the FactCC metric showing lower scores than the BART model generated summaries, especially in the more abstractive XSum setting. The DAE metric for the baseline model is slightly lower than the BART model scores in the CNN/DM setting and has no improvement in the XSum setting showing that it does not perform corrections on complex errors. These results confirm our hypothesis that the baseline model trained on adversarial data based on heuristic errors does not transfer well to real errors in model generated summaries. In contrast, our model improves both metrics across both datasets. On the more challenging XSum dataset, our model has a $\sim$17 point improvement on FactCC and $\sim$0.1 improvement on DAE over the baseline model. The BART generated summaries on CNN/DM are $\sim$70\% factual and hence using the FactCC Filter to correct only non-factual summaries helps improve results on FactCC.  As XSum has more than 80\% non-factual summaries, the FactCC filter does not change results and correcting all generated summaries is beneficial. In Table \ref{tab:qual} we present examples of corrections made by \ourmodel{} and present a discussion in \Sref{sec:appendix_qual}.

Prior works have shown that improving factual consistency in summaries leads to a drop in ROUGE scores \cite{factentailment,cao2021cliff, cao-etal-2020-factual}. Our
ROUGE results do not drop significantly and are consistent with prior work. These results show that our model does not significantly change the summaries and the corrected summaries contain the relevant information w.r.t.~to the source. 
\begin{figure*}[t]
    {\includegraphics[width=\textwidth]{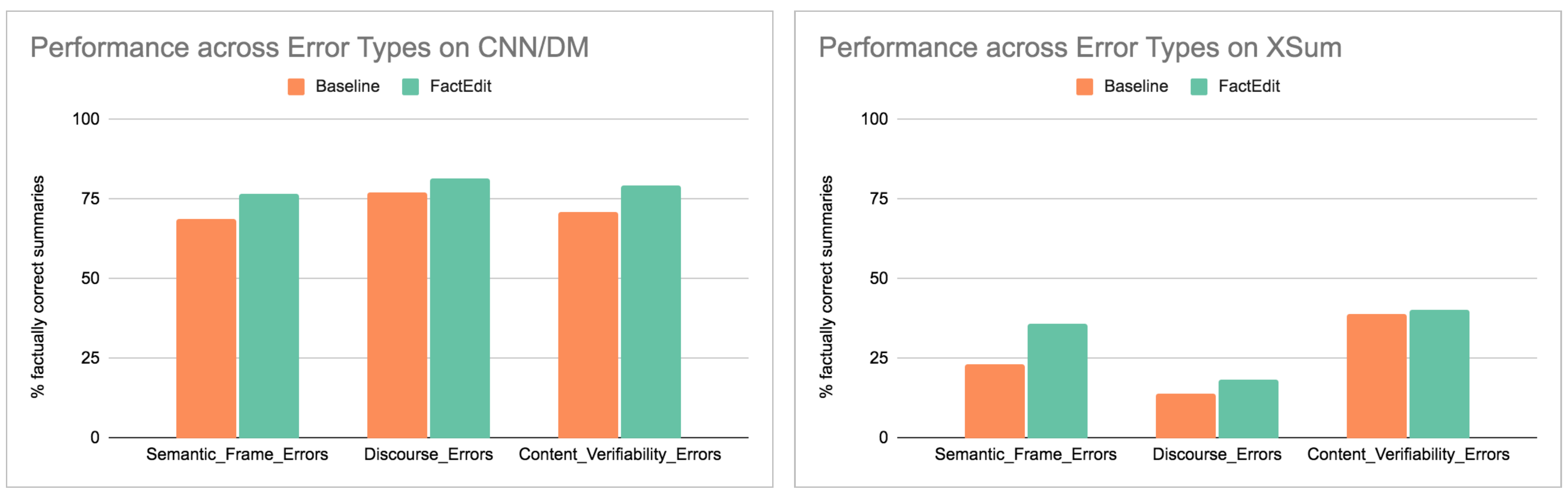}}
    \caption{Performance of FactEdit across different error categories in comparison to baseline \cite{cao-etal-2020-factual}}. \ourmodel{} improves the percentage of factual summaries across diverse types of factual errors.
    \label{fig:err_type_res}
\vspace{-0.5em}
\end{figure*}

\subsection{Factuality Results across Model Types}
\label{sec:res_model}
Table \ref{tab:frank_res} shows results of using \ourmodel{} to correct summaries generated by different types of language models using the FRANK benchmark \cite{frank}. We provide the generated summaries collected in the benchmark along with their source document as input to our trained fact corrector. This setting evaluates the generalizability of our adversarial training data in handling different error distributions from different summarization models. Our results show our model significantly improves the factuality in generated summaries across 8 out of 9 test models. The FactCC Filter helps improves results in CNN/DM setting but does not change results in XSum similar to results in \Sref{sec:res_fact}. In the more extractive CNN/DM setting, fact correction improves FactCC scores by $\sim$5.3 points and DAE scores by $\sim$10.9 points on average across models. In the more challenging and abstractive XSum dataset, we improve FactCC scores by $\sim$24 points and DAE scores by $\sim$31 points on average. Our results show that our model trained using Infilling LM based adversarial data is able to generalize and correct errors in generated summaries across different model types. Further, the significant improvement in XSum suggests that using LMs to generate factually incorrect candidates produces rich negative examples which help correct errors in more abstractive summaries.

Pretrained models like BART, BertSum and BertS2S have improved generation capabilities and make lesser mistakes in generating the right entity or predicate and more mistakes in discourse structuring \cite{frank}. \ourmodel{} correspondingly shows larger improvements in DAE scores than FactCC scores in these pretrained models. The Pointer-Generator model being highly extractive in nature scores highly in factuality metrics in the CNN/DM setting and \ourmodel{}  reduces results in this setting showing that our model is not beneficial in copy-based model settings. On the other hand, in the XSum setting, the base Pointer-Generator model scores poorly and correcting factual errors in them improves factuality scores. Non-pretrained sequence-to-sequence models like Seq2Seq and TransSeq2Seq score poorly in both ROUGE and Factuality scores due to their limited language generation capabilities. By correcting factual errors in them, we improve factuality metrics significantly without changes in ROUGE, indicating that the gains are due to fact correction and not just rewriting the summary using a strong language model.

\section{Analysis}
\label{sec:analysis}
\subsection{Performance across Error Categories}
\label{sec:res_err}
The FRANK benchmark proposes a typology of three coarse categories of error types and collects human annotations on the error category: i) Semantic Frame Errors - This category covers factual errors in a sentence due to incorrect entity or predicate being generated ii) Discourse Errors - This covers discourse level factual errors like incorrect pronouns or sentence ordering iii) Content Verifiability Errors - This category is for errors whose factuality cannot be judged either due to grammatical errors or 
hallucinated content. We evaluate our model on its ability to correct different types of errors. We use the generated summaries from the best pretrained model in FRANK for each dataset - BART for CNN/DM and BertS2S for XSum. For each subset of summaries of a particular error type, we correct the summaries using \ourmodel{} and report the percentage of factual summaries in the output as predicted by Ent-DAE. We compare \ourmodel{} with the baseline to study whether our model improves error correction for each type. 

From \fref{fig:err_type_res}, we see that across both datasets \ourmodel{} increases the percentage of factual summaries across all three error categories, showing that the data generation process in \ourmodel{} can generalize across multiple error types without the need for special heuristics or annotations. We see the largest improvements in the Semantic Frame Error category with an increase of $\sim$ 8 points on CNN/DM and $\sim$ 13 points on XSum. On the more complex Discourse Errors we see an improvement of $\sim$ 5 points on both datasets. Finally, on Content Verifiability Errors, we see a $\sim$ 8 point improvement on CNN/DM and $\sim$ 2 point improvement on XSum. XSum has a high proportion of hallucination errors
and our results highlight the challenge in correcting this error type.

\subsection{Transferrability across Datasets}
\label{sec:res_data}
\begin{table}[t]
\centering
\small
\begin{tabular}{c|c|c}
 \textbf{Method} & \textbf{FactCC} & \textbf{Ent-DAE}\\
 \hline
 BertS2S & 22.29 & 05.71 \\
  \ourmodel{} (FF) - CNN Model & \textbf{33.71} & \textbf{22.29}\\
 \hline
 TConvS2S & 17.71 & 04.00 \\
 \ourmodel{} (FF) - CNN Model & \textbf{30.29} & \textbf{22.29}\\
 \cline{1-3}
 PointGen & 18.29 & 00.57 \\
 \ourmodel{} (FF) - CNN Model & \textbf{28.57} & \textbf{19.43}\\
 \cline{1-3}
 TranS2S & 18.86 & 2.86 \\
 \ourmodel{} (FF) - CNN Model & \textbf{18.86} & \textbf{21.14}\\
\end{tabular}
\caption{Transfer results of \ourmodel{}. \ourmodel{} trained using CNN/DM data transfers well to summaries generated for documents in XSum.
}
\vspace{-1.5em}
\label{tab:frank_res_tranf}
\end{table}

It is not always feasible to train specialized fact correction models for each dataset or style of summaries. While CNN/DM and XSum contain documents of the news domain, they both have different summary characteristics. Certain applications might benefit from a single model which can generalize to different summary styles. We evaluate the ability of \ourmodel{} trained on CNN/DM data (\ourmodel{} FF - CNN Model) to transfer and correct summaries generated for XSum documents using FRANK benchmark. Table \ref{tab:frank_res_tranf} shows results for this experiment.
Our results show significant improvement in factuality scores across all model types in this setting, showing that our data generation process produces rich and diverse factually incorrect examples which can generalize to factual errors in other data settings. By using only the source documents, our training data is agnostic of the styles, lengths and characteristics of reference summaries and hence is able to generalize to the headline style abstractive summaries of XSum.

\begin{table}[t]
\centering
\small
\begin{tabular}{l|c|c}
\textbf{Method} & \textbf{Fluency} & \textbf{Factuality}\\
\hline
\citet{cao-etal-2020-factual} & 4.58 & 3.10 \\
\ourmodel{} & 4.75 & 3.33 \\
\end{tabular}
\caption{Results of Human Evaluation on Fluency and Factuality of corrected summaries. Human judges rate summaries corrected by \ourmodel{} 
higher in fluency and factuality than the baseline.}
\vspace{-1.5em}
\label{tab:ablation}
\end{table}

\subsection{Human Evaluation}
To further study whether the factuality corrections performed by our model align with human expectations of automated summaries, we conduct a human study. Two annotators evaluated 20 randomly sampled summaries generated from the test set of the XSum dataset using the BertS2S model and corrected by \ourmodel{} and the baseline. The annotators were shown the entire source document and one corrected summary at a time and asked to rate the fluency and factuality of the summary on a 1-5 Likert scale. In manual evaluation,
annotators rated \ourmodel{} an average of 3.3 on factuality and 4.8 on fluency, compared to the baseline which was rated 3.1 and 4.6 scores respectively, showing that \ourmodel{} improves on both factuality and fluency.

\subsection{Ablation Study}
\label{sec:res_abl}

Our model corrects each sentence in a summary given context of the rest of the summary and relevant passages in the source document. We ablate this setup by removing parts of the context one at a time. In Table \ref{tab:ablation} we present the results. We observe a a drop in results when using the entire summary as context (-RelevPass) and when removing the context of the summary in which the sentence occurs (-SummCtxt). Our results show the importance of having the appropriate context to enable the model to perform fact correction well.


\begin{table}[t]
\centering
\small
\begin{tabular}{l|c|c}
\textbf{Method} & \textbf{FactCC} & \textbf{E-DAE}\\
\hline
\multicolumn{3}{c}{CNN/DM}\\
\hline
\ourmodel{} & 76.03 & 75.36 \\
\ourmodel{} -SummCtxt & 75.73 & 74.23 \\
\ourmodel{} -SummCtxt-RelevPass & 75.89 & 75.03 \\
\hline 
\multicolumn{3}{c}{Xsum}\\
\hline
\ourmodel{} & 23.91 & 20.13 \\
\ourmodel{} -SummCtxt & 22.89 & 20.06 \\
\ourmodel{} -SummCtxt-RelevPass & 23.48 & 20.08 \\
\end{tabular}
\caption{Results of Ablation study with components of fact correction pipeline removed. SummCtxt includes the generated summary as additional context. RelevPass includes relevant passages from the source as additional context. \ourmodel{} setup 
ourperforms the ablated versions on FactCC and DAE scores.}
\label{tab:ablation}
\end{table}

\section{Related Work}
\label{sec:related_work}

\noindent\textbf{Factuality Evaluation}
Standard n-gram based metrics do not correlate well with human judgements of factuality and are unsuitable for evaluating factuality \cite{sumcriticaleval, fabbri2020summeval}. Several automated metrics were proposed to detect factual errors in generated summaries. They primarily fall in two paradigms---Entailment based and QA based metrics. \citet{goodrich19,factcc,factentailment, entdae} model factuality as an entailment verifying whether the summary is entailed by the source. \citet{lee2022masked} use similar masked infilling to generate training data for such metrics.
QA models can be used to answer questions about the document, separately using the article and the output summary as context and compare the answers to score the factuality of summaries \cite{feqa, factqa}.
To evaluate these metrics, recent work collec human judgements for factuality \cite{fabbri2020summeval, factentailment, frank}. Additionally, \citep{frank} also obtain annotations on factual error categories, which we use for our evaluations.
This paper considers the problem of improving factuality, not measuring it. While this is a different task, it is related: e.g., measuring the number of corrections made by \ourmodel{} might be useful as a factuality measure.

\noindent\textbf{Improving Factuality of Summaries:} There are two paradigms of work to ensure generated summaries are factually consistent: i) imposing factuality constraints during training or generation and ii) post-editing generated summaries to correct factual errors. \citet{wan2022factpegasus} add factuality contraints during pretraining by using factually consistent summaries. Model designs and factuality specific objectives help optimize for factuality during training \cite{gabriel2019discourse, cao2021cliff, dong2022faithful, rajagopal2022counterfactual}.
During decoding beam search candidates can be ranked based on factuality measures \cite{king2022don, zhao2020reducing}. Work on correcting factual errors post generation is relatively nascent. \citet{cao-etal-2020-factual} and \citet{lee2022factual} train fact correction models on synthetic data based on heuristic errors which we show is less effective than LM based error generation (\Tref{tab:overall_res}).
\citet{dong-etal-2020-multi} use a QA model to replace phrases in the summary with spans in the source text. This requires multiple inference iterations, making them very expensive for correction. In contrast our approach corrects errors in one iteration, making it a faster and more practical approach for error correction. Tangentially, work on correcting errors in reference summaries to make the training data more reliable has also been explored \citet{adams2022learning, wan2022factpegasus}. In dialog generation, \citet{Gupta2021SynthesizingAN} explore using mask-fill approaches to generate synthetic data for response ranking, showing that using language models to generate adversarial data might be applicable beyond summarization.

\section*{Conclusion}
We present an adversarial data generation process to generate rich synthetic data for a post editing model, which can be applied to correct factual errors generated summaries. Our data generation process leverages Infilling Language Models to produce alternative candidate summaries. Using the generated data, we train models to rewrite summaries by correcting factual errors in them. Through extensive experiments across two datasets and nine models, we show that our fact corrector model improves the factual consistency of the summaries, making them more reliable. 


\section*{Limitations}
Our model is trained to rewrite generated summaries by correcting factual errors in them. A limitation in our current setup is accurate detection of factual errors. We rely on off-the-shelf metrics for identifying summaries with factual errors to correct. Our model does not perform detection and correction together and often rewrites correct summaries as well if fed to the model. Therefore for settings like CNN/DM, it's beneficial to filter summaries using a factuality metric before giving summaries to our model as input. As our fact corrector is a sequence-to-sequence model, it could potentially introduce new factual errors in the summaries. It is essential to use factually detectors to ensure summaries are factual before real world usage of any corrected summary.

\section*{Ethical Considerations}
State-of-the-art language generation models, including summarization, are not yet powerful enough to facilitate fine-grained control over generated content. 
This leads to problems with content fidelity and safety; our work aims to ameliorate issues related to factual reliability of the models. 
However, existing approaches, including ours, cannot guarantee this yet. 
Furthermore, there is a risk of dual use, since the same techniques can be used to post-edit models to produce non-factual, harmful content to mislead, impersonate, or manipulate opinions. Future research should focus on developing better defenses methods against mis-using language generators maliciously.  

\section*{Acknowledgements}
We would like to thank Lucille Njoo, Xiaochuang Han, Sachin Kumar, Dheeraj Rajagopal and other members of the Tsvetshop Lab for their valuable feedback on this work. This material is based upon work supported by the DARPA CMO under Contract No.~HR001120C0124. Any opinions, findings and conclusions or recommendations expressed in this material are those of the author(s) and do not necessarily state or reflect those of the United States Government or any agency thereof. 
Y.T.~also gratefully acknowledges support from NSF CAREER Grant No.~IIS2142739 and Alfred P. Sloan Foundation Fellowship.
\bibliography{anthology,custom}

\begin{thebibliography}{41}
\expandafter\ifx\csname natexlab\endcsname\relax\def\natexlab#1{#1}\fi

\bibitem[{Adams et~al.(2022)Adams, Shing, Sun, Winestock, McKeown, and
  Elhadad}]{adams2022learning}
Griffin Adams, Han-Chin Shing, Qing Sun, Christopher Winestock, Kathleen
  McKeown, and No{\'e}mie Elhadad. 2022.
\newblock Learning to revise references for faithful summarization.
\newblock \emph{arXiv preprint arXiv:2204.10290}.

\bibitem[{Banko et~al.(2007)Banko, Cafarella, Soderland, Broadhead, and
  Etzioni}]{openie}
Michele Banko, Michael~J. Cafarella, Stephen Soderland, Matt Broadhead, and
  Oren Etzioni. 2007.
\newblock \href {http://dl.acm.org/citation.cfm?id=1625275.1625705} {Open
  information extraction from the web}.
\newblock In \emph{Proceedings of the 20th International Joint Conference on
  Artifical Intelligence}, IJCAI'07, pages 2670--2676, San Francisco, CA, USA.
  Morgan Kaufmann Publishers Inc.

\bibitem[{Cao et~al.(2020)Cao, Dong, Wu, and Cheung}]{cao-etal-2020-factual}
Meng Cao, Yue Dong, Jiapeng Wu, and Jackie Chi~Kit Cheung. 2020.
\newblock \href {https://doi.org/10.18653/v1/2020.emnlp-main.506} {Factual
  error correction for abstractive summarization models}.
\newblock In \emph{Proceedings of the 2020 Conference on Empirical Methods in
  Natural Language Processing (EMNLP)}, pages 6251--6258, Online. Association
  for Computational Linguistics.

\bibitem[{Cao and Wang(2021)}]{cao2021cliff}
Shuyang Cao and Lu~Wang. 2021.
\newblock Cliff: Contrastive learning for improving faithfulness and factuality
  in abstractive summarization.
\newblock \emph{arXiv preprint arXiv:2109.09209}.

\bibitem[{Cao et~al.(2018)Cao, Wei, Li, and Li}]{fact-aware}
Ziqiang Cao, Furu Wei, Wenjie Li, and Sujian Li. 2018.
\newblock \href
  {https://www.aaai.org/ocs/index.php/AAAI/AAAI18/paper/view/16121} {Faithful
  to the original: Fact aware neural abstractive summarization}.
\newblock In \emph{Proceedings of the Thirty-Second {AAAI} Conference on
  Artificial Intelligence, (AAAI-18), the 30th innovative Applications of
  Artificial Intelligence (IAAI-18), and the 8th {AAAI} Symposium on
  Educational Advances in Artificial Intelligence (EAAI-18), New Orleans,
  Louisiana, USA, February 2-7, 2018}, pages 4784--4791. {AAAI} Press.

\bibitem[{Devlin et~al.(2019)Devlin, Chang, Lee, and Toutanova}]{bert}
Jacob Devlin, Ming-Wei Chang, Kenton Lee, and Kristina Toutanova. 2019.
\newblock \href {https://doi.org/10.18653/v1/N19-1423} {{BERT}: Pre-training of
  deep bidirectional transformers for language understanding}.
\newblock In \emph{Proceedings of the 2019 Conference of the North {A}merican
  Chapter of the Association for Computational Linguistics: Human Language
  Technologies, Volume 1 (Long and Short Papers)}, pages 4171--4186,
  Minneapolis, Minnesota. Association for Computational Linguistics.

\bibitem[{Donahue et~al.(2020)Donahue, Lee, and Liang}]{donahue2020enabling}
Chris Donahue, Mina Lee, and Percy Liang. 2020.
\newblock Enabling language models to fill in the blanks.
\newblock \emph{arXiv preprint arXiv:2005.05339}.

\bibitem[{Dong et~al.(2020)Dong, Wang, Gan, Cheng, Cheung, and
  Liu}]{dong-etal-2020-multi}
Yue Dong, Shuohang Wang, Zhe Gan, Yu~Cheng, Jackie Chi~Kit Cheung, and Jingjing
  Liu. 2020.
\newblock \href {https://doi.org/10.18653/v1/2020.emnlp-main.749} {Multi-fact
  correction in abstractive text summarization}.
\newblock In \emph{Proceedings of the 2020 Conference on Empirical Methods in
  Natural Language Processing (EMNLP)}, pages 9320--9331, Online. Association
  for Computational Linguistics.

\bibitem[{Dong et~al.(2022)Dong, Wieting, and Verga}]{dong2022faithful}
Yue Dong, John Wieting, and Pat Verga. 2022.
\newblock Faithful to the document or to the world? mitigating hallucinations
  via entity-linked knowledge in abstractive summarization.
\newblock \emph{arXiv preprint arXiv:2204.13761}.

\bibitem[{Dreyer et~al.(2021)Dreyer, Liu, Nan, Atluri, and
  Ravi}]{dreyer2021analyzing}
Markus Dreyer, Mengwen Liu, Feng Nan, Sandeep Atluri, and Sujith Ravi. 2021.
\newblock Analyzing the abstractiveness-factuality tradeoff with nonlinear
  abstractiveness constraints.
\newblock \emph{arXiv preprint arXiv:2108.02859}.

\bibitem[{Durmus et~al.(2020)Durmus, He, and Diab}]{feqa}
Esin Durmus, He~He, and Mona Diab. 2020.
\newblock \href {https://doi.org/10.18653/v1/2020.acl-main.454} {{FEQA}: A
  question answering evaluation framework for faithfulness assessment in
  abstractive summarization}.
\newblock In \emph{Proceedings of the 58th Annual Meeting of the Association
  for Computational Linguistics}, pages 5055--5070, Online. Association for
  Computational Linguistics.

\bibitem[{Fabbri et~al.(2020)Fabbri, Kry{\'s}ci{\'n}ski, McCann, Xiong, Socher,
  and Radev}]{fabbri2020summeval}
Alexander~R Fabbri, Wojciech Kry{\'s}ci{\'n}ski, Bryan McCann, Caiming Xiong,
  Richard Socher, and Dragomir Radev. 2020.
\newblock Summeval: Re-evaluating summarization evaluation.
\newblock \emph{arXiv preprint arXiv:2007.12626}.

\bibitem[{Gabriel et~al.(2019)Gabriel, Bosselut, Da, Holtzman, Buys, Lo,
  Celikyilmaz, and Choi}]{gabriel2019discourse}
Saadia Gabriel, Antoine Bosselut, Jeff Da, Ari Holtzman, Jan Buys, Kyle Lo,
  Asli Celikyilmaz, and Yejin Choi. 2019.
\newblock Discourse understanding and factual consistency in abstractive
  summarization.
\newblock \emph{arXiv preprint arXiv:1907.01272}.

\bibitem[{Gehrmann et~al.(2018)Gehrmann, Deng, and Rush}]{bus}
Sebastian Gehrmann, Yuntian Deng, and Alexander Rush. 2018.
\newblock \href {https://doi.org/10.18653/v1/D18-1443} {Bottom-up abstractive
  summarization}.
\newblock In \emph{Proceedings of the 2018 Conference on Empirical Methods in
  Natural Language Processing}, pages 4098--4109, Brussels, Belgium.
  Association for Computational Linguistics.

\bibitem[{Goodrich et~al.(2019)Goodrich, Rao, Liu, and Saleh}]{goodrich19}
Ben Goodrich, Vinay Rao, Peter~J. Liu, and Mohammad Saleh. 2019.
\newblock \href {https://doi.org/10.1145/3292500.3330955} {Assessing the
  factual accuracy of generated text}.
\newblock In \emph{Proceedings of the 25th {ACM} {SIGKDD} International
  Conference on Knowledge Discovery {\&} Data Mining, {KDD} 2019, Anchorage,
  AK, USA, August 4-8, 2019}, pages 166--175. {ACM}.

\bibitem[{Goyal and Durrett(2020)}]{dae}
Tanya Goyal and Greg Durrett. 2020.
\newblock \href {https://doi.org/10.18653/v1/2020.findings-emnlp.322}
  {Evaluating factuality in generation with dependency-level entailment}.
\newblock In \emph{Findings of the Association for Computational Linguistics:
  EMNLP 2020}, pages 3592--3603, Online. Association for Computational
  Linguistics.

\bibitem[{Goyal and Durrett(2021)}]{entdae}
Tanya Goyal and Greg Durrett. 2021.
\newblock \href {https://doi.org/10.18653/v1/2021.naacl-main.114} {Annotating
  and modeling fine-grained factuality in summarization}.
\newblock In \emph{Proceedings of the 2021 Conference of the North American
  Chapter of the Association for Computational Linguistics: Human Language
  Technologies}, pages 1449--1462, Online. Association for Computational
  Linguistics.

\bibitem[{Gupta et~al.(2021)Gupta, Tsvetkov, and
  Bigham}]{Gupta2021SynthesizingAN}
Prakhar Gupta, Yulia Tsvetkov, and Jeffrey~P. Bigham. 2021.
\newblock Synthesizing adversarial negative responses for robust response
  ranking and evaluation.
\newblock In \emph{FINDINGS}.

\bibitem[{Hermann et~al.(2015)Hermann, Kocisk{\'{y}}, Grefenstette, Espeholt,
  Kay, Suleyman, and Blunsom}]{cnn-dm}
Karl~Moritz Hermann, Tom{\'{a}}s Kocisk{\'{y}}, Edward Grefenstette, Lasse
  Espeholt, Will Kay, Mustafa Suleyman, and Phil Blunsom. 2015.
\newblock \href
  {https://proceedings.neurips.cc/paper/2015/hash/afdec7005cc9f14302cd0474fd0f3c96-Abstract.html}
  {Teaching machines to read and comprehend}.
\newblock In \emph{Proceedings of the Advances in Neural Information Processing
  Systems 28: Annual Conference on Neural Information Processing Systems 2015,
  December 7-12, 2015, Montreal, Quebec, Canada}, pages 1693--1701.

\bibitem[{Huang et~al.(2021)Huang, Feng, Feng, and Qin}]{huang2021factual}
Yichong Huang, Xiachong Feng, Xiaocheng Feng, and Bing Qin. 2021.
\newblock The factual inconsistency problem in abstractive text summarization:
  A survey.
\newblock \emph{arXiv preprint arXiv:2104.14839}.

\bibitem[{Hutson et~al.(2021)}]{hutson2021robo}
Matthew Hutson et~al. 2021.
\newblock Robo-writers: the rise and risks of language-generating ai.
\newblock \emph{Nature}, 591(7848):22--25.

\bibitem[{King et~al.(2022)King, Shen, Subramani, Weld, Beltagy, and
  Downey}]{king2022don}
Daniel King, Zejiang Shen, Nishant Subramani, Daniel~S Weld, Iz~Beltagy, and
  Doug Downey. 2022.
\newblock Don't say what you don't know: Improving the consistency of
  abstractive summarization by constraining beam search.
\newblock \emph{arXiv preprint arXiv:2203.08436}.

\bibitem[{Kryscinski et~al.(2019)Kryscinski, Keskar, McCann, Xiong, and
  Socher}]{sumcriticaleval}
Wojciech Kryscinski, Nitish~Shirish Keskar, Bryan McCann, Caiming Xiong, and
  Richard Socher. 2019.
\newblock \href {https://doi.org/10.18653/v1/D19-1051} {Neural text
  summarization: A critical evaluation}.
\newblock In \emph{Proceedings of the 2019 Conference on Empirical Methods in
  Natural Language Processing and the 9th International Joint Conference on
  Natural Language Processing (EMNLP-IJCNLP)}, pages 540--551, Hong Kong,
  China. Association for Computational Linguistics.

\bibitem[{Kryscinski et~al.(2020)Kryscinski, McCann, Xiong, and
  Socher}]{factcc}
Wojciech Kryscinski, Bryan McCann, Caiming Xiong, and Richard Socher. 2020.
\newblock \href {https://doi.org/10.18653/v1/2020.emnlp-main.750} {Evaluating
  the factual consistency of abstractive text summarization}.
\newblock In \emph{Proceedings of the 2020 Conference on Empirical Methods in
  Natural Language Processing (EMNLP)}, pages 9332--9346, Online. Association
  for Computational Linguistics.

\bibitem[{Lee et~al.(2022{\natexlab{a}})Lee, Park, Yoon, Bui, Dernoncourt, Kim,
  and Jung}]{lee2022factual}
Hwanhee Lee, Cheoneum Park, Seunghyun Yoon, Trung Bui, Franck Dernoncourt, Juae
  Kim, and Kyomin Jung. 2022{\natexlab{a}}.
\newblock Factual error correction for abstractive summaries using entity
  retrieval.
\newblock \emph{arXiv preprint arXiv:2204.08263}.

\bibitem[{Lee et~al.(2022{\natexlab{b}})Lee, Yoo, Park, Lee, and
  Jung}]{lee2022masked}
Hwanhee Lee, Kang~Min Yoo, Joonsuk Park, Hwaran Lee, and Kyomin Jung.
  2022{\natexlab{b}}.
\newblock Masked summarization to generate factually inconsistent summaries for
  improved factual consistency checking.
\newblock \emph{arXiv preprint arXiv:2205.02035}.

\bibitem[{Lewis et~al.(2020)Lewis, Liu, Goyal, Ghazvininejad, Mohamed, Levy,
  Stoyanov, and Zettlemoyer}]{bart}
Mike Lewis, Yinhan Liu, Naman Goyal, Marjan Ghazvininejad, Abdelrahman Mohamed,
  Omer Levy, Veselin Stoyanov, and Luke Zettlemoyer. 2020.
\newblock \href {https://doi.org/10.18653/v1/2020.acl-main.703} {{BART}:
  Denoising sequence-to-sequence pre-training for natural language generation,
  translation, and comprehension}.
\newblock In \emph{Proceedings of the 58th Annual Meeting of the Association
  for Computational Linguistics}, pages 7871--7880, Online. Association for
  Computational Linguistics.

\bibitem[{Lin(2004)}]{lin2004rouge}
Chin-Yew Lin. 2004.
\newblock Rouge: A package for automatic evaluation of summaries.
\newblock In \emph{Text summarization branches out}, pages 74--81.

\bibitem[{Liu and Lapata(2019)}]{liu-lapata-2019-text}
Yang Liu and Mirella Lapata. 2019.
\newblock \href {https://doi.org/10.18653/v1/D19-1387} {Text summarization with
  pretrained encoders}.
\newblock In \emph{Proceedings of the 2019 Conference on Empirical Methods in
  Natural Language Processing and the 9th International Joint Conference on
  Natural Language Processing (EMNLP-IJCNLP)}, pages 3730--3740, Hong Kong,
  China. Association for Computational Linguistics.

\bibitem[{Maynez et~al.(2020)Maynez, Narayan, Bohnet, and
  McDonald}]{factentailment}
Joshua Maynez, Shashi Narayan, Bernd Bohnet, and Ryan McDonald. 2020.
\newblock \href {https://doi.org/10.18653/v1/2020.acl-main.173} {On
  faithfulness and factuality in abstractive summarization}.
\newblock In \emph{Proceedings of the 58th Annual Meeting of the Association
  for Computational Linguistics}, pages 1906--1919, Online. Association for
  Computational Linguistics.

\bibitem[{Narayan et~al.(2018)Narayan, Cohen, and Lapata}]{xsum}
Shashi Narayan, Shay~B. Cohen, and Mirella Lapata. 2018.
\newblock \href {https://doi.org/10.18653/v1/D18-1206} {Don{'}t give me the
  details, just the summary! topic-aware convolutional neural networks for
  extreme summarization}.
\newblock In \emph{Proceedings of the 2018 Conference on Empirical Methods in
  Natural Language Processing}, pages 1797--1807, Brussels, Belgium.
  Association for Computational Linguistics.

\bibitem[{Pagnoni et~al.(2021)Pagnoni, Balachandran, and Tsvetkov}]{frank}
Artidoro Pagnoni, Vidhisha Balachandran, and Yulia Tsvetkov. 2021.
\newblock \href {https://doi.org/10.18653/v1/2021.naacl-main.383}
  {Understanding factuality in abstractive summarization with {FRANK}: A
  benchmark for factuality metrics}.
\newblock In \emph{Proceedings of the 2021 Conference of the North American
  Chapter of the Association for Computational Linguistics: Human Language
  Technologies}, pages 4812--4829, Online. Association for Computational
  Linguistics.

\bibitem[{Rajagopal et~al.(2022)Rajagopal, Shakeri, Santos, Hovy, and
  Chang}]{rajagopal2022counterfactual}
Dheeraj Rajagopal, Siamak Shakeri, Cicero Nogueira~dos Santos, Eduard Hovy, and
  Chung-Ching Chang. 2022.
\newblock Counterfactual data augmentation improves factuality of abstractive
  summarization.
\newblock \emph{arXiv preprint arXiv:2205.12416}.

\bibitem[{Ranade et~al.(2021)Ranade, Joshi, and Finin}]{aigenfakenews}
Priyanka Ranade, Anupam Joshi, and Tim Finin. 2021.
\newblock \href
  {https://theconversation.com/study-shows-ai-generated-fake-reports-fool-experts-160909}
  {Study shows ai-generated fake reports fool experts}.

\bibitem[{Rush et~al.(2015)Rush, Chopra, and Weston}]{rush}
Alexander~M. Rush, Sumit Chopra, and Jason Weston. 2015.
\newblock \href {https://doi.org/10.18653/v1/D15-1044} {A neural attention
  model for abstractive sentence summarization}.
\newblock In \emph{Proceedings of the 2015 Conference on Empirical Methods in
  Natural Language Processing}, pages 379--389, Lisbon, Portugal. Association
  for Computational Linguistics.

\bibitem[{See et~al.(2017)See, Liu, and Manning}]{pgn}
Abigail See, Peter~J. Liu, and Christopher~D. Manning. 2017.
\newblock \href {https://doi.org/10.18653/v1/P17-1099} {Get to the point:
  Summarization with pointer-generator networks}.
\newblock In \emph{Proceedings of the 55th Annual Meeting of the Association
  for Computational Linguistics (Volume 1: Long Papers)}, pages 1073--1083,
  Vancouver, Canada. Association for Computational Linguistics.

\bibitem[{Vaswani et~al.(2017)Vaswani, Shazeer, Parmar, Uszkoreit, Jones,
  Gomez, Kaiser, and Polosukhin}]{vaswani2017attention}
Ashish Vaswani, Noam Shazeer, Niki Parmar, Jakob Uszkoreit, Llion Jones,
  Aidan~N. Gomez, Lukasz Kaiser, and Illia Polosukhin. 2017.
\newblock \href
  {https://proceedings.neurips.cc/paper/2017/hash/3f5ee243547dee91fbd053c1c4a845aa-Abstract.html}
  {Attention is all you need}.
\newblock In \emph{Proceedings of the Advances in Neural Information Processing
  Systems 30: Annual Conference on Neural Information Processing Systems 2017,
  December 4-9, 2017, Long Beach, CA, {USA}}, pages 5998--6008.

\bibitem[{Wan and Bansal(2022)}]{wan2022factpegasus}
David Wan and Mohit Bansal. 2022.
\newblock Factpegasus: Factuality-aware pre-training and fine-tuning for
  abstractive summarization.
\newblock \emph{arXiv preprint arXiv:2205.07830}.

\bibitem[{Wang et~al.(2020)Wang, Cho, and Lewis}]{factqa}
Alex Wang, Kyunghyun Cho, and Mike Lewis. 2020.
\newblock \href {https://doi.org/10.18653/v1/2020.acl-main.450} {Asking and
  answering questions to evaluate the factual consistency of summaries}.
\newblock In \emph{Proceedings of the 58th Annual Meeting of the Association
  for Computational Linguistics}, pages 5008--5020, Online. Association for
  Computational Linguistics.

\bibitem[{Zhang et~al.(2020)Zhang, Zhao, Saleh, and Liu}]{zhang2020pegasus}
Jingqing Zhang, Yao Zhao, Mohammad Saleh, and Peter Liu. 2020.
\newblock Pegasus: Pre-training with extracted gap-sentences for abstractive
  summarization.
\newblock In \emph{International Conference on Machine Learning}, pages
  11328--11339. PMLR.

\bibitem[{Zhao et~al.(2020)Zhao, Cohen, and Webber}]{zhao2020reducing}
Zheng Zhao, Shay~B Cohen, and Bonnie Webber. 2020.
\newblock Reducing quantity hallucinations in abstractive summarization.
\newblock \emph{arXiv preprint arXiv:2009.13312}.

\end{thebibliography}
\bibliographystyle{acl_natbib}

\appendix

\section{Qualitative Analysis}
\label{sec:appendix_qual}

\begin{table*}[t]
\centering
\resizebox{1.0\textwidth}{!}{%
\begin{tabular}{@{}lll@{}}
\toprule
Source & \begin{tabular}[c]{@{}l@{}} Generated Summary \end{tabular} & \begin{tabular}[c]{@{}l@{}} Corrected Summary \end{tabular} \\ 
\midrule
\begin{tabular}[c]{@{}l@{}}An endangered bottlenose dolphin has died off the coast of Spain\\It had lost its mother and become stranded in shallow waters\\ off the coast of Mojacar in Andalucia. Tourists then began touching\\ and photographing the animal rather than seeking help for it.\\  Equinac, a local NGO, said "selfishness" caused the dolphin\\ "suffering and stress."  The organisation, which works to protect \\marine wildlife in the area, wrote in a Facebook post (in Spanish):\\ "The animal was subjected to the curious who wanted to photograph\\ and touch it."  "These animals are highly protected; to disturb them, to \\harm them, to manipulate them and to harass them is prohibited\\ by law, and we always ask for respect and consideration."\end{tabular} & \begin{tabular}[c]{@{}l@{}}An endangered bottlenose dolphin has died\\ off the coast of Spain after being harassed by tourists\\ who wanted to take pictures of it, an animal welfare \\group has said in a Facebook post, in part because \\of their "selfish" behaviour.774,000 people have\\ visited the animal since it was stranded.\end{tabular} & \begin{tabular}[c]{@{}l@{}}An endangered bottlenose dolphin has died \\off the coast of Spain after being harassed by tourists \\who wanted to take pictures of it, an animal welfare \\group has said in a Facebook post, in part because\\ of their "selfish" behaviour.\end{tabular} \\ \midrule
\begin{tabular}[c]{@{}l@{}}BBC Wales Sport understands the Gills made an enquiry for Byrne\\, before Edinburgh changed his mind over a player he brought\\ to Rodney Parade.  Edinburgh opted to end the pursuit out of respect\\ for his old club, who he managed from 2011-2015, so as not to weaken\\ their battle against relegation.  They are currently 20th in League Two,\\ seven points above the relegation zone.  Meanwhile, Newport\\ have confirmed the release of defender Matt Taylor.  The 34-year\\ old joined from Cheltenham in the summer\\ but made just five appearances for the south Wales club.\\  "Further to recent discussions, we can confirm that Matt Taylor\\ will leave the Club with immediate effect having mutually\\ agreed to terminate his contract," a statement from the club read.\\ \end{tabular} & \begin{tabular}[c]{@{}l@{}}Newport County manager Justin Edinburgh has ended\\ his interest in signing Gillingham midfielder Luke Byrne,\\ BBC Wales Sport has learned, after talks between the two\\ clubs ended in stalemate on Tuesday \\night, reports BBC Radio Wales' Mark McGhee\\ and BBC Radio Newport's Steve Davies.\end{tabular} & \begin{tabular}[c]{@{}l@{}}Newport County manager Justin Edinburgh has ended\\ his interest in signing Gillingham midfielder Luke Byrne,\\ BBC Wales Sport has learned, after talks between the two\\ clubs ended in stalemate on Tuesday night,\\ reports BBC Wales Sport.\end{tabular} \\ \bottomrule
\end{tabular}%
}
\caption{Examples of corrections made by our fact corrector.}
\label{tab:qual}
\end{table*}

In Table \ref{tab:qual} we present examples of corrections made by our models. In the first example, the generated summary contains an entire sentence which is unsupported by the source document. \ourmodel{} chooses to remove the entire sentence instead of rewriting or correcting it. In the second example, the generated summary contains hallucinated entities of reporter names which are not present in the source. Here, the \ourmodel{} rewrites by replacing the incorrect entity with the correct ones.

\end{document}